\documentclass{article}

\usepackage[final]{neurips_2023}

\usepackage{color}
\definecolor{darkblue}{rgb}{0.0,0.0,0.5}

\usepackage[utf8]{inputenc} 
\usepackage[T1]{fontenc}    
\usepackage[colorlinks=true,linkcolor=darkblue,urlcolor=darkblue,citecolor=darkblue,anchorcolor=blue]{hyperref}       
\usepackage{url}            
\usepackage{booktabs}       
\usepackage{amsfonts}       
\usepackage{nicefrac}       
\usepackage{microtype}      
\usepackage{xcolor}         

\title{Neither hype nor gloom do DNNs justice\\\vspace{0.25cm}
  \large A commentary on \cite{bowers2022deep}}

\author{Felix A.\ Wichmann$^1$, Simon Kornblith$^2$, Robert Geirhos$^2$}
\date{$^1$ Neural Information Processing Group, University of T\"{u}bingen, Germany\\$^2$Google Research, Brain Team, Toronto, Canada}

\author{%
  Felix A.\ Wichmann\\
  Neural Information Processing Group\\
  University of T\"{u}bingen\\
  \And
  Simon Kornblith\\
  Work done at Google Research, Brain Team\\
  Now at Anthropic\\
  \And
  Robert Geirhos\\
  Work done at Google Research, Brain Team\\
  Now at Google DeepMind\\
}

\begin{document}

\maketitle

\begin{abstract}
Neither the hype exemplified in some exaggerated claims about deep neural networks (DNNs), nor the gloom expressed by Bowers et al. do DNNs as models in vision science justice: DNNs rapidly evolve, and today's limitations are often tomorrow's successes. In addition, providing explanations as well as prediction and image-computability are model desiderata; one should not be favoured at the expense of the other.
\end{abstract}

We agree with \cite{bowers2022deep} that some of the quoted statements at the beginning of their target article about DNNs as ``best models'' are exaggerated---perhaps some of them bordering on scientific hype \citep{Intemann_2020}. However, only the authors of such exaggerated statements are to blame, not DNNs: Instead of blaming DNNs, perhaps Bowers et al.\ should have engaged in a critical discussion of the increasingly widespread practice of rewarding impact and boldness over carefulness and modesty that allows hyperbole to flourish in science.
This is unfortunate as the target article does mention a number of valid issues with DNNs in vision science and raises a number of valid concerns. For example, we fully agree that human vision is much more than recognising photographs of objects in scenes; we also fully agree there are still a number of important behavioural differences between DNNs and humans even in terms of core object recognition \citep{DiCarlo_2012}, i.e.\ even when recognising photographs of objects in scenes, such as DNNs' adversarial susceptibility (section 4.1.1) or reliance on local rather than global features (section 4.1.3). However, we do not subscribe to the somewhat gloomy view of DNNs in vision science expressed by Bowers et al. We believe that image-computable models are essential to the future of vision science, and DNNs are currently the most promising---albeit not yet fully adequate---model class for core object recognition.

Importantly, any behavioural differences between DNNs and humans can only be a snapshot in time---true as of today.  Unlike Bowers et al.\ we do not see any evidence that future, novel DNN architectures, training data and regimes may not be able to overcome at least some of the limitations mentioned in the target article---and Bowers et al.\ certainly do not provide any convincing evidence why solving such tasks is beyond DNNs \emph{in principle, i.e.\ forever}. In just over a decade, DNNs have come a long way from AlexNet, and we still witness tremendous progress in deep learning. Until recently, DNNs lacked robustness to image distortions; now some match or outperform humans on many of them. DNNs made very different error patterns than humans; newer models achieve at least somewhat better consistency \citep{geirhos2021partial}. DNNs used to be texture-biased; now some are shape-biased similar to humans \citep{dehghani2023scaling}. With DNNs, today's limitations are often tomorrow's success stories. 

Yes, current DNNs fail on a large number of ``psychological tasks'', from (un-)crowding \citep{Doerig-etal_2020b} to focusing on local rather than global shape \citep{Baker-etal_2018}, from similarity judgements \citep{German-Jacobs_2020} to combinatorial judgements \citep{Montero-etal_2022}; furthermore, current DNNs lack (proper, human-like) sensitivity to Gestalt principles \citep{Biscione-Bowers_2023}. But current DNNs in vision are typically trained to recognize static images; their failure on ``psychological tasks'' without (perhaps radically) different training or different optimisation objectives does not surprise us---just as we do not expect a traditional vision model of motion processing to predict lightness induction or an early spatial vision model to predict Gestalt laws, at least not without substantial modification and fitting it to suitable data. To overcome current DNNs' limitations on psychological tasks we need more DNN research inspired by vision science, not just engineering to improve models' overall accuracy---here we certainly agree again with Bowers et al.

Moreover, for many of the above mentioned psychological tasks, there simply do not exist successful traditional vision models. Why single out DNNs as failures if no successful computational model exists, at least not image-computable models? Traditional ``object recognition'' models only model isolated aspects of object recognition, and it is difficult to tell how well they model these aspects, since only image-computable models can actually recognize objects. Here, image-computability is far more than just a ``nice to have'' criterion since it facilitates falsifiability. Bowers et al.'s long list of DNN failures should rather be taken as a list of desiderata of what future image-computable models of human vision should explain and predict.

Although we do not know whether DNNs will be sufficient to meet this challenge, only future research will resolve the many open questions: Is our current approach of applying predominantly discriminative DNNs as computational models of human vision sufficient to obtain truly successful models? Do we need to incorporate, for example, causality \citep{Pearl_2009a}, or generative models such as predictive coding \citep{Rao_1999} or even symbolic computations \citep{mao2018the}? Do we need to ground learning in intuitive theories of physics and psychology \citep{Lake_2017}?

Finally, it appears as if Bowers et al.\ argue that models should first and foremost provide explanations, as if predictivity---which includes but is not limited to image-computability---did not matter much.\footnote{Or observational data; successful models need to be able to explain and predict data from hypothesis driven experiments as well as observational data.} While we agree with Bowers et al.\ that in machine learning there is a tendency to blindly chase improved benchmark numbers without seeking understanding of underlying phenomena, we believe that both prediction and explanation are required: an explanation without prediction cannot be trusted, and a prediction without explanation does not aid understanding. What we need is not a myopic focus on one or the other, but to be more explicit about modelling goals---in the target article by Bowers et al.\ and in general as we argue in a forthcoming article \citep{Wichmann-Geirhos_2023}.

We think that neither the hype exemplified in some exaggerated claims about DNNs, nor the gloom expressed by Bowers et al.\ do DNNs and their application to vision science justice. Looking forward, if we want to make progress towards modelling and understanding human visual perception, we believe that it will be key to move beyond both hype and gloom and carefully explore similarities and differences between human vision and rapidly evolving DNNs.

\paragraph{Funding statement} This work was supported by the German Research Foundation (DFG): SFB 1233, Robust
Vision: Inference Principles and Neural Mechanisms, TP 4, project number: 276693517 to F.A.W. In addition, F.A.W. is a member of the Machine Learning Cluster of Excellence, funded by the Deutsche Forschungsgemeinschaft (DFG, German Research Foundation) under Germany’s Excellence Strategy – EXC number 2064/1 – Project number 390727645.

\bibliographystyle{apalike}
\bibliography{refs}

\end{document}